\documentclass[11pt]{article}

\usepackage[preprint]{acl}

\usepackage{times}
\usepackage{latexsym}

\usepackage[T1]{fontenc}

\usepackage[utf8]{inputenc}

\usepackage{microtype}

\usepackage{inconsolata}

\usepackage{graphicx}
\usepackage{bm, amsfonts, amsmath}
\usepackage{booktabs}
\usepackage{multirow}
\usepackage{tabularx}
\usepackage{array}
\usepackage{microtype}
\usepackage{algpseudocode}
\usepackage{enumitem}
\usepackage{subcaption}
\usepackage{mathtools}
\usepackage{comment}
\usepackage{xcolor}

\title{\textit{Hallucinate at the Last} in Long Response Generation: A Case Study on Long Document Summarization}

\author{
Joonho Yang$^{1}$,~Seunghyun Yoon$^{2}$,~Hwan Chang$^{1}$,~Byeongjeong Kim$^{1}$,~Hwanhee Lee$^{1}$\thanks{Corresponding author.} \\
    $^{1}$Department of Artificial Intelligence, Chung-Ang University, $^{2}$Adobe Research, USA\\
    \texttt{\{plm3332, hwanchang, michael97k, hwanheelee\}@cau.ac.kr}, \texttt{syoon@adobe.com}
}

\begin{document}
\maketitle
\renewcommand*{\thefootnote}{\arabic{footnote}}
\begin{abstract}
Large Language Models (LLMs) have significantly advanced text generation capabilities, including tasks like summarization, often producing coherent and fluent outputs.
However, faithfulness to the source material remains a significant challenge due to the generation of hallucinations by LLMs.
While extensive research focuses on detecting and reducing these inaccuracies, less attention has been paid to the positional distribution of hallucinations within generated text, particularly in long outputs.
In this work, we investigate \textit{where} hallucinations frequently occur in LLM-based long response generation, using long document summarization as a key case study.
Focusing on the challenging setting of long context-aware long response generation, we find a consistent and concerning phenomenon: hallucinations tend to concentrate disproportionately in the latter parts of the generated long response.
To understand this positional bias, we explore potential contributing factors related to the dynamics of attention and decoding over long sequences.
Furthermore, we investigate methods to mitigate this positional hallucination, aiming to improve faithfulness specifically in the concluding segments of long outputs.

\end{abstract}
\section{Introduction}
Recent advancements in Large Language Models (LLMs) have pushed the boundaries of human-like language generation, particularly in tasks like text summarization~\cite{chang2024survey}.
With their enhanced scale and sophisticated training, LLMs now produce summaries exhibiting remarkable coherence and fluency, approaching human-level quality~\cite{song-etal-2025-learning}.
This capability is transforming how we interact with large volumes of text, making information more accessible~\cite{openai2024gpt4technicalreport, geminiteam2024gemini15unlockingmultimodal}.

\begin{figure}[!t]
\centering
\includegraphics[width=0.95\linewidth]{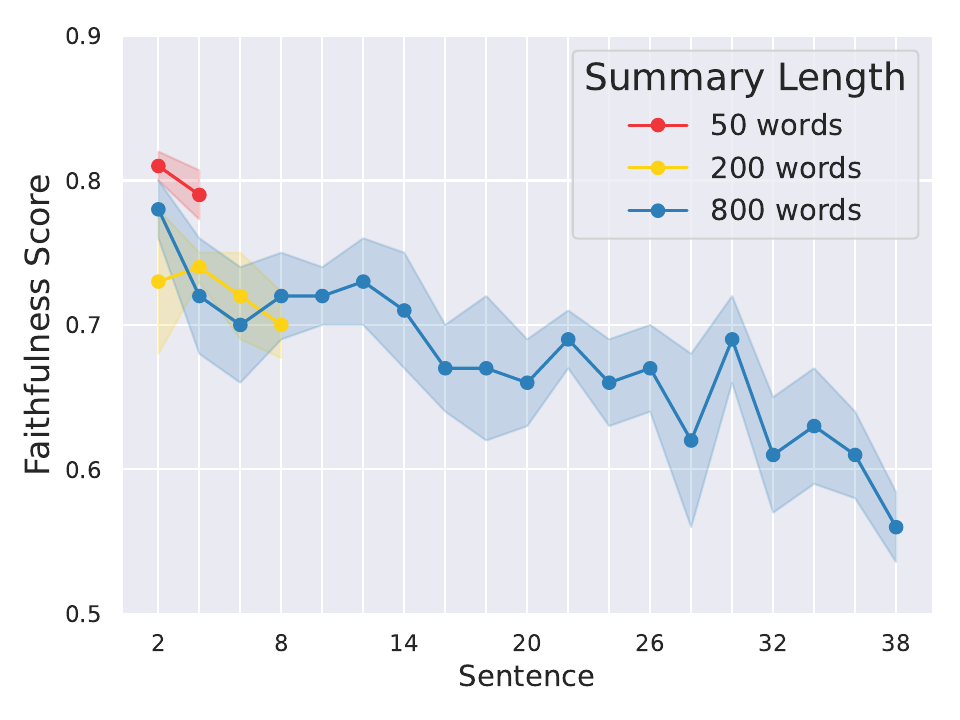}
\caption{Comparison of the faithfulness scores of summaries generated around 50, 200, and 800 words using GPT-4o mini.}
\label{fig1}
\vspace{-5mm}
\end{figure} 


As LLMs continue to evolve, a paradigm shift is emerging from short-form to long-form generation, enabled by the ability to process extended contexts~\cite{wu2025shiftinglongcontextllmsresearch}.
Long-output generation is critical for complex reasoning tasks like long Chain-of-Thought prompting~\cite{jaech2024openai} and language agents~\cite{sumers2024cognitivearchitectureslanguageagents, Wang_2024}, inherently requiring coherent, contextually grounded long responses.
While recent benchmarks evaluate long-generation capabilities~\cite{wu2025longgenbench, ye2025longprocbenchmarkinglongcontextlanguage}, they often lack contextual grounding, overlooking a crucial aspect of long-form generation: \textit{context-aware faithfulness}.

A persistent and critical challenge faced by LLMs is the phenomenon of \textit{hallucination}, wherein the generated content is unfaithful to or unsupported by the input context~\cite{Huang_2025}.
While extensive prior work has focused on detecting and mitigating hallucinations, a significant limitation is the lack of research into the positional distribution of factual errors within the generated sequence. Related research has explored positional biases in how models process \textit{input contexts}, notably the \textit{Lost in the Middle} phenomenon~\cite{liu-etal-2024-lost, wan-etal-2025-positional}. However, understanding the distribution of errors \textit{within the generated output itself}, particularly for long responses from long contexts, is equally crucial for effective diagnosis and mitigation.

Motivated by this critical research gap, our work provides the first dedicated investigation into the positional distribution of hallucinations during \textit{long-response generation}, especially in the challenging setting of long document summarization. Unlike prior work primarily addressing long input contexts or short outputs~\cite{zhang-etal-2024-fine, bishop-etal-2024-longdocfactscore}, we tackle \textit{long context-aware long response generation}, which requires processing lengthy inputs and maintaining faithfulness over a significantly long output sequence. Specifically, we pose a core research question: How frequently do hallucinations occur in long-form generation tasks such as document summarization, and when they do, where in the output are they most likely to appear?

Our analysis reveals a surprising and concerning trend: hallucinations tend to concentrate towards the end of the generated text, a phenomenon we term "hallucinate at the last."
As shown in Figure~\ref{fig1}, faithfulness significantly decreases towards the end for long summaries (e.g., 800 words).
Contrary to expectations of uniform distribution, this distinct positional bias highlights a critical vulnerability in LLMs when generating extended text.
To systematically analyze this phenomenon and its implications, our research addresses three core questions:

\begin{enumerate}[wide, labelwidth=!, label={\textbf{RQ\arabic*.}}, labelindent=0pt, topsep=2pt, itemsep=-1pt, itemindent=0pt, leftmargin=*, before=\setlength{\listparindent}{-\leftmargin}]
\item Where Do Hallucinations Most Frequently Occur? (\S\ref{rq1})
\item What Factors Contribute to Hallucination Concentration at the Last Part? (\S\ref{rq2})
\item How Can We Mitigate the Hallucination in the Last Part? (\S\ref{rq3})
\end{enumerate}

To answer these questions, we empirically characterize the positional distribution of hallucinations (\textbf{RQ1}), investigate their generative causes (\textbf{RQ2}), and propose mitigation strategies (\textbf{RQ3}).
Our results underscore the importance of the positional dimension of errors in building robust long-form generation systems.

Our contributions can be summarized as follows:
\begin{itemize}[leftmargin=10pt, labelindent=0pt]
\item We provide the first empirical characterization of the \textit{Hallucinate at the Last} phenomenon in LLM-based long response generation, particularly in long document summarization.

\item We offer initial insights and analysis into the potential factors contributing to this late-stage hallucination.
\item We investigate mitigation strategies specifically tailored to address positional hallucination.
\end{itemize}

\section{Generating Summaries}
\label{sec:generating_summaries}

This section describes the experimental setup used to generate the summaries analyzed in our study.
We prompt LLMs to generate summaries from original documents.
The prompt is provided in Appendix~\ref{sec:appx_generating_summaries}.



\subsection{Input Context Lengths}
We evaluate model performance with input length of over 4K tokens.
For the Llama model specifically, we further extend the evaluation to context lengths of 12K and 16K tokens.

\subsection{Output Length Categories}
We define two categories for output length to study the \textit{Hallucinate at the Last} phenomenon:
\begin{itemize}[leftmargin=*,itemsep=0.1em]
    \item \textbf{Standard Summary:} Summaries with a length between 100 and 200 words.
    \footnote{The length of the standard summaries in our study was determined based on the average length of the reference summaries in the dataset, which is employed in Section~\ref{rq2}.}
    \item \textbf{Long Summary:} Summaries with a length up to 30\% of the input context length.
\end{itemize}
This distinction allows us to compare hallucination patterns in standard summary lengths versus significantly longer generated responses.
By including both standard and long output categories, we aim to systematically analyze if and how the positional bias of hallucinations manifests and potentially becomes more severe as the length of the generated summary increases. 
See Appendix~\ref{sec:appx_generating_summaries} for more details.

\subsection{Overall Faithfulness} 

As a preliminary step, we measure the overall faithfulness of the generated long summaries with FineSurE~\cite{song-etal-2024-finesure} to ensure consistent quality over various context lengths, prior to our main positional analysis.
Table~\ref{tab:tab1} shows results for Llama3.1-8B-Instruct~\cite{grattafiori2024llama3herdmodels} on the Wikipedia dataset, evaluated using FineSurE with GPT-4~\cite{openai2024gpt4technicalreport}.
FineSurE assesses each sentence against the source, identifying error types when inconsistencies occur.
The results align with those on CNNDM~\cite{10.5555/2969239.2969428}, where Llama3-70B-Instruct achieved 85.5\% faithfulness.
We next conduct a finer-grained positional analysis in Section~\ref{rq1} for various setups.
\section{Where Do Hallucinations Most Frequently Occur?}
\label{rq1}
In this section, we empirically investigate where hallucinations most frequently occur in long document summarization. We analyze the positional distribution of factual errors by examining faithfulness across varying models, datasets, and output positions.


\begin{table}[!t]
    \centering
    \small
    \resizebox{0.7\linewidth}{!}{
        \begin{tabular}{c|c}
            \toprule
            \textbf{\textsc{Context Length}} & \textbf{\textsc{Faithfulness(\%)}} \\
            \midrule
            4K & 82.9 \\
            5K & 86.6 \\
            6K & 85.9 \\
            7K & 83.9 \\
            8K & 86.8 \\
            \bottomrule
        \end{tabular}
    }
    \caption{Overall faithfulness of \textbf{long} summaries on the Wikipedia dataset, evaluated using FineSurE.}
\label{tab:tab1}
\vspace{-5mm}
\end{table}

\subsection{Experimental Setup}
\label{ssec:experimental_setup}
\paragraph{Models}
We evaluate our approach on multiple state-of-the-art LLMs, including Llama3.1-8B-Instruct, Gemma3-12B-Instruct~\cite{team2025gemma}, Qwen2.5-7B-Instruct~\cite{qwen2025qwen25technicalreport}, and GPT-4o mini, to ensure model-agnostic findings.
For Gemma, we set the sliding window attention context length to 1 token (as it cannot be disabled entirely), whereas for Qwen, we modify the default setting of over 131K tokens to 16K tokens.
Throughout all experiments, summaries are generated using greedy decoding.

\paragraph{Datasets}
To examine whether the phenomenon is domain-specific, we analyze the positional distribution of faithfulness within generated summaries across various domains.
In addition to the Wikipedia dataset used in the main experiment, we include the arXiv~\cite{cohan-etal-2018-discourse}, PubMed~\cite{cohan-etal-2018-discourse}, and GovReport~\cite{huang-etal-2021-efficient} datasets for further evaluation.
For each dataset, we randomly sampled 20 examples for analysis.

\subsection{Evaluation Metric}
\label{ssec:evaluation_metric}
\paragraph{Faithfulness}
To enable an in-depth analysis of positional bias within each generated summary, we use the MiniCheck-Flan-T5-Large~\cite{tang-etal-2024-minicheck} model to evaluate faithfulness.
MiniCheck is particularly suitable for this fine-grained analysis, as it is an efficient and high-performing fact-checking model that leverages atomic facts~\cite{min-etal-2023-factscore, jing-etal-2024-faithscore, yu2024rlaifvopensourceaifeedback}.

Let $s_i$ denote the $i$-th sentence in a generated summary $S$.
Each sentence is first decomposed into a set of atomic facts $A_i$ using an LLM, followed by a filtering step to remove unnecessary atomic facts~\cite{yang-etal-2024-fizz}, which makes:
\begin{equation}
A_i = \{a_{i1}, a_{i2}, ..., a_{iN_i}\}
\label{eq:eq1}
\end{equation}
We then compare each filtered atomic fact with the source document $D={d_1, ..., d_M}$ using the MiniCheck model $\mathcal{M}$ in a pairwise fashion.
For each atomic fact, we take the maximum score obtained across all source sentences:
\begin{equation}
\textrm{score}(a_{ij}) = \max_{d_{m}\in{D}}\mathcal{M}(d_{m}, a_{ij})
\label{eq:eq2}
\end{equation}
Finally, the sentence-level \textbf{faithfulness score} is computed as the average score of all filtered atomic facts within that sentence:
\begin{equation}
\textrm{Faithfulness}(s_i) = \frac{1}{N_{i}}\sum_{j=1}^{N_i}\textrm{score}(a_{ij})
\label{eq:eq3}
\end{equation}
Each sentence is assigned a faithfulness score in the range $[0,1]$, where 1 indicates that all atomic facts in the sentence are fully supported by the source document, and 0 indicates a complete lack of support.

\paragraph{Sensitivity}
In addition to the faithfulness score, we propose a simple yet effective metric for analyzing positional discrepancies in generated outputs.
Following~\citet{singh-goshtasbpour-2022-platt}, each summary is divided into five equal length bins, and the average faithfulness score of the atomic facts within each bin is computed.
The bins are defined using the \texttt{cut} function in the pandas package, and the scores of the atomic facts are assigned to bins using open-interval boundaries, meaning that boundary values are not included.
Let $f_k$ denote the average faithfulness score of the $k$-th bin
($k=1,\dots,5$). We define \textbf{sensitivity} as:
\[
\text{Sensitivity}
= 100 \times 
\left(
\frac{1}{4}\sum_{k=1}^{4} f_k-f_5
\right).
\]
A positive value indicates that the faithfulness in the final bin is lower than in the earlier bins (i.e., a tendency to hallucinate at the end), whereas a negative value suggests the opposite trend.
Larger sensitivity values correspond to stronger hallucination effects near the end of the generated output.

Additional experimental details and an example of the evaluation procedure are provided in Appendix~\ref{sec:appx_evaluation_metric}.
Furthermore, the human evaluation of the quality and completeness of the generated atomic facts and experimental details is presented in Appendix~\ref{sec:appx_atomic_fact}.

\subsection{Hallucinate at the Last}
\begin{figure}[t]
\centering
\includegraphics[width=1.0\linewidth]{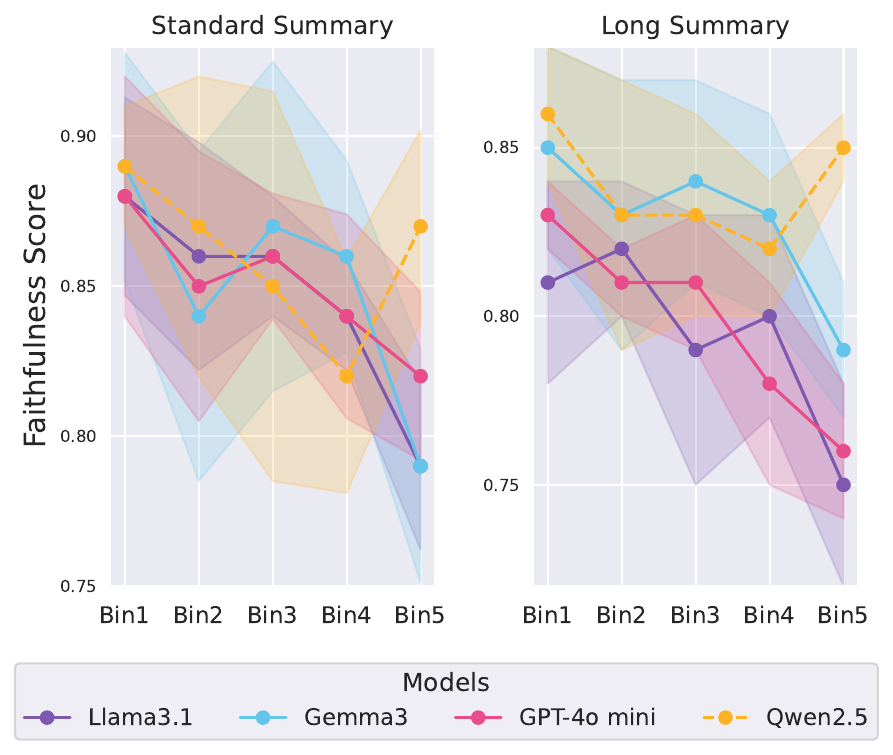}
\caption{Comparison of faithfulness scores for summaries generated by different models on the Wikipedia dataset.} 
\label{fig2}
\vspace{-5mm}
\end{figure}

\paragraph{The latter part of the summary exhibits the lowest level of faithfulness across multiple models, consistently observed across summary lengths. The decline becomes most pronounced when the summaries are long.}
We present the positional faithfulness discrepancies of both standard and long summaries generated by different models on the Wikipedia dataset in Figure~\ref{fig2}.
As shown in the figure, both standard and long summaries produced by the three models, excluding Qwen, consistently exhibit a decline in faithfulness toward the later portions of the summary, with the lowest scores typically appearing at the end.
This \textit{Hallucinate at the Last} phenomenon becomes even more pronounced in long summaries: as the summary length increases, the faithfulness near the end continues to decrease, eventually dropping below 0.75 for these models.
In contrast, Qwen displays a different pattern.
Its faithfulness tends to drop most sharply in the middle of the summary, after which the scores recover toward the end.

\begin{table}[!b]
    \centering
    \resizebox{\linewidth}{!}{
        \begin{tabular}{c|ccccc|c}
            \toprule
            \multirow{2}{*}{\textbf{\textsc{Models}}} & \multicolumn{5}{c|}{\textbf{\textsc{Generated Summary Bins}}} & \multirow{2}{*}{\textbf{\textsc{Sensitivity}}} \\
            & \textbf{\textsc{1}} & \textbf{\textsc{2}} & \textbf{\textsc{3}} & \textbf{\textsc{4}} & \textbf{\textsc{5}} & \\
            \midrule
            Llama3.1-8B-Instruct & 0.81 & \textbf{0.82} & 0.79 & 0.80 & \underline{0.75} & 5.50 \\
            \addlinespace
            Llama3.1-70B-Instruct & \textbf{0.83} & 0.82 & \textbf{0.83} & 0.81 & \underline{0.77} & 5.25 \\
            \midrule
            \addlinespace
            Gemma3-12B-Instruct & \textbf{0.85} & 0.83 & 0.84 & 0.83 & \underline{0.79} & 4.75 \\
            \midrule
            GPT-4o mini & \textbf{0.83} & 0.81 & 0.81 & 0.78 & \underline{0.76} & 4.75 \\
            \midrule
            Qwen2.5-7B-Instruct & \textbf{0.86} & 0.83 & 0.83 & \underline{0.82} & 0.84 & -0.50 \\
            \bottomrule
        \end{tabular}
    }
    \caption{Faithfulness scores and sensitivity for \textbf{long} summaries generated by various models on the Wikipedia dataset.}
\label{tab:bins_all}
\vspace{-5mm}
\end{table}

\begin{table}[!t]
    \centering
    \resizebox{\linewidth}{!}{
        \begin{tabular}{c|c|ccccc|c}
            \toprule
            \multirow{2}{*}{\textbf{\textsc{Models}}} & \textbf{\textsc{Context}} & \multicolumn{5}{c|}{\textbf{\textsc{Generated Summary Bins}}} & \multirow{2}{*}{\textbf{\textsc{Sensitivity}}} \\
            & \textbf{\textsc{Length}} & \textbf{\textsc{1}} & \textbf{\textsc{2}} & \textbf{\textsc{3}} & \textbf{\textsc{4}} & \textbf{\textsc{5}} & \\
            \midrule
            \multicolumn{8}{c}{\textbf{Standard Summary}} \\
            \midrule
            \multirow{2}{*}{Llama3.1-8B-Instruct} & 12K & \textbf{0.91} & 0.87 & 0.85 & 0.76 & \underline{0.67} & 17.65 \\
            & 16K & \textbf{0.84} & 0.83 & 0.76 & 0.71 & \underline{0.69} & 3.00 \\
            \midrule
            \multicolumn{8}{c}{\textbf{Long Summary}} \\
            \midrule
            \multirow{2}{*}{Llama3.1-8B-Instruct} & 12K & \textbf{0.80} & 0.79 & 0.66 & \underline{0.63} & 0.69 & 9.50 \\
            & 16K & \textbf{0.86} & 0.83 & 0.85 & 0.83 & \underline{0.79} & 5.25 \\
            \bottomrule
        \end{tabular}
    }
    \caption{Faithfulness scores and sensitivity for \textbf{standard} and \textbf{long} summaries generated by Llama on the Wikipedia dataset, with 12K and 16K context length.}
\label{tab:bins_llama_1216}
\vspace{-3mm}
\end{table}
\begin{table}[!t]
    \centering
    \resizebox{\linewidth}{!}{
        \begin{tabular}{c|ccccc|c}
            \toprule
            \multirow{2}{*}{\textbf{\textsc{Model}}} & \multicolumn{5}{c|}{\textbf{\textsc{Generated Summary Bins}}} & \multirow{2}{*}{\textbf{\textsc{Sensitivity}}} \\
            & \textbf{\textsc{1}} & \textbf{\textsc{2}} & \textbf{\textsc{3}} & \textbf{\textsc{4}} & \textbf{\textsc{5}} & \\
            \midrule
            Llama3.1-8B-Instruct & \textbf{0.90} & 0.87 & 0.84 & 0.84 & \underline{0.70} & 16.25 \\
            \bottomrule
        \end{tabular}
    }
    \caption{Human evaluation of faithfulness scores and sensitivity for \textbf{long} summaries generated by Llama on the Wikipedia dataset.}
\label{tab:bins_llama_human}
\vspace{-5mm}
\end{table}


\begin{figure}[!b]
\centering
\includegraphics[width=1.0\linewidth]{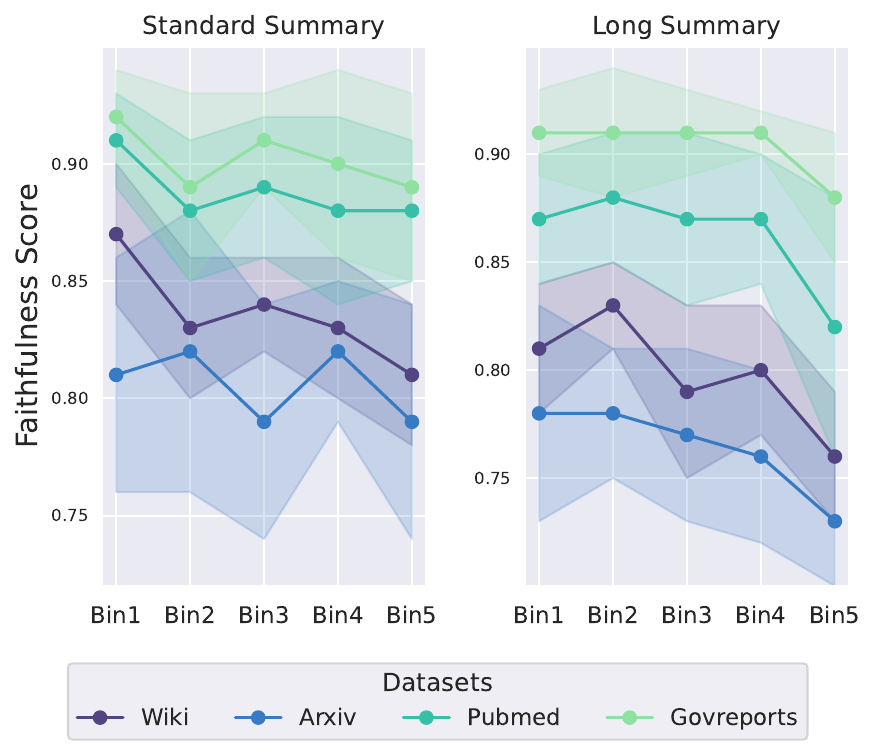}
\caption{Comparison of faithfulness scores for summaries generated by the Llama3.1-8B-Instruct model in multiple domains.
}
\label{fig4}
\vspace{-5mm}
\end{figure}

Table~\ref{tab:bins_all} report the faithfulness and sensitivity values for the different models.
All models except Qwen show markedly low faithfulness scores in the final bin, indicating high sensitivity.
Consistent with the human evaluation in Table~\ref{tab:bins_llama_human}, Llama demonstrates the highest sensitivity regardless of model size.
In contrast, Qwen is the only model that records negative sensitivity values.
Unlike the consistent decline observed in the other models, Qwen maintains relatively stable faithfulness across bins, resulting in negative sensitivity.
We provide a detailed analysis of these contrasting patterns in Section~\ref{rq2}.

Table~\ref{tab:bins_llama_1216} presents the faithfulness and sensitivity results for both the standard and long summaries generated by the LLaMA model with context lengths of 12K and 16K tokens.
Interestingly, we find that positional discrepancies in the generated summaries are not significantly affected by the input context length.
As shown in the table, even with longer contexts of approximately 12K and 16K tokens, sensitivity remains high, consistently demonstrating the \textit{Hallucinate at the Last} phenomenon.
Supplementary results across different context lengths and methods are presented in Appendix~\ref{sec:appx_different_methods}.


\paragraph{The latter part of the summary consistently demonstrates the lowest level of faithfulness across all datasets, with a marked decline on the long summaries.}
We present the positional faithfulness discrepancy of long summaries generated by the Llama model across different datasets in Figure~\ref{fig4}.
As the results indicate, the \textit{Hallucinate at the Last} tendency is consistently observed across multiple domains.
Faithfulness scores gradually decrease toward the final segments as the output length increases, and this decline becomes most severe in the concluding portions of longer summaries.
Additional sensitivity results across diverse datasets are provided in Appendix~\ref{sec:appx_varying_datasets}.

\paragraph{The \textit{Hallucinate at the Last} phenomenon persists across different decoding strategies.}
To demonstrate that the \textit{Hallucinate at the Last} phenomenon is not merely a consequence of error propagation, we first conduct experiments using various decoding strategies.
Following ~\citet{shi-etal-2024-thorough}, we adopt the following decoding strategies to generate summaries.

\begin{itemize}[leftmargin=*,itemsep=0.1em]
    \item \textbf{Temperature \& Top-\textit{p} Sampling} samples tokens from the next-token probability distribution using the smallest set of tokens whose cumulative probability mass exceeds a given threshold~\textit{p}.
    We conduct experiments with temperature values $\tau \in \{0.5, 0.7\}$ and with top-\textit{p} thresholds of 0.7 and 0.9.
    \item \textbf{Top-\textit{k} Sampling} samples tokens from the \textit{k} most probable candidates.
    We experiment with \textit{k} values of 20 and 100.
    \item \textbf{$\eta$-Sampling} discards tokens whose probabilities fall below an entropy dependent threshold.
    The hyperparameter $\eta$ is explored with 6e-4 and 4e-3.
\end{itemize}

\begin{figure}[t]
\centering
\includegraphics[width=1.0\linewidth]{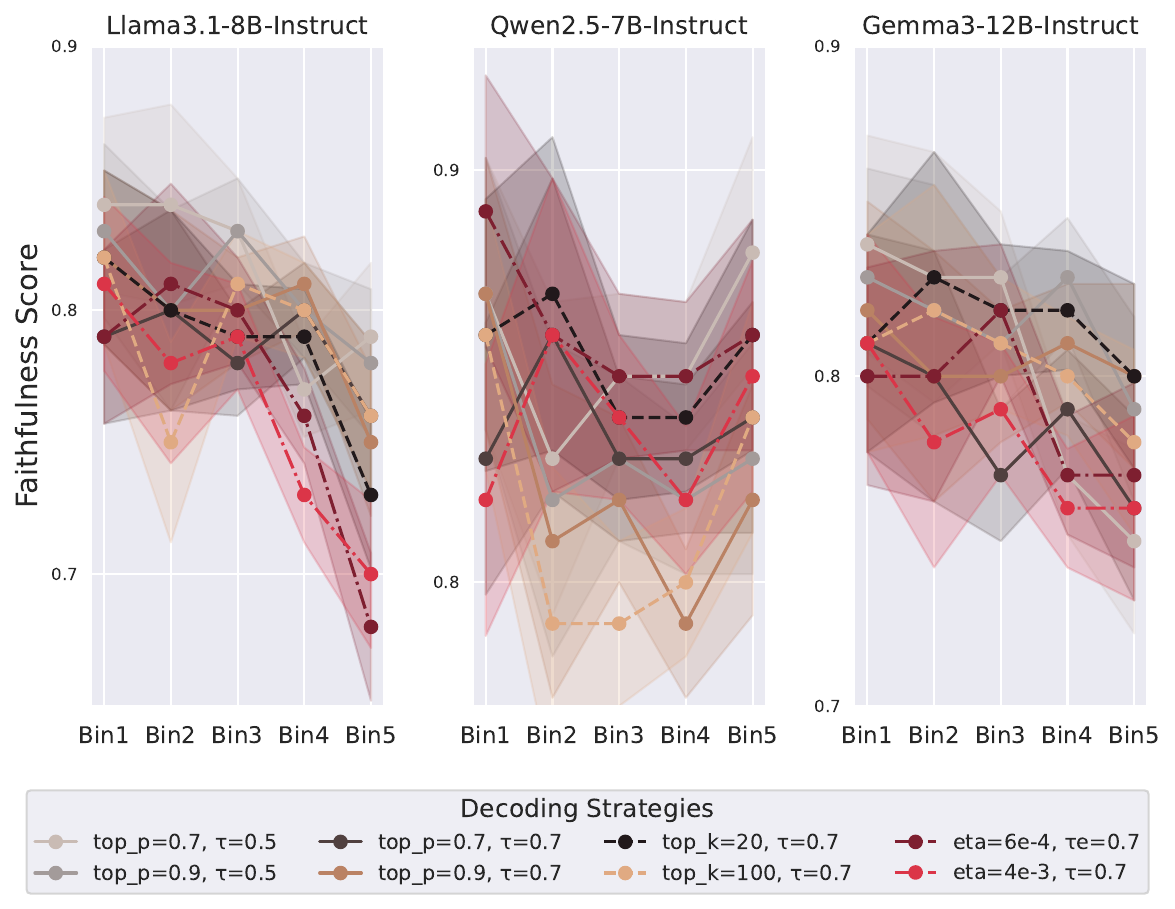}
\caption{Comparison of faithfulness scores for \textbf{long} summaries generated using different decoding strategies on the Wikipedia dataset.} 
\label{fig_decoding}
\vspace{-5mm}
\end{figure}

We present the positional faithfulness discrepancies in long summaries generated using different decoding strategies in Figure~\ref{fig_decoding}.
For the Llama and Gemma models, the latter portions of the summaries exhibit the lowest levels of faithfulness across all decoding strategies.
This effect is particularly pronounced under $\eta$-sampling for Llama, where the faithfulness score remains above 0.8 in the earlier segments but drops below 0.7 in the final bin.
In contrast, the Qwen model shows a different trend.
With the exception of two decoding strategies, the faithfulness of the generated summaries does not decline toward the end but instead increases.
Notably, the faithfulness scores for Qwen are lowest in the middle portions of the summaries rather than at the end.
Motivated by these observations, we further investigate this phenomenon across a broader set of models.

\paragraph{Human Evaluation}
Each atomic fact was labeled as true or false based on whether it was supported by the source document.
True labels were scored as 1 and false as 0, and these scores were averaged within the same five positional bins as in the automatic evaluation.
The resulting human-judged faithfulness scores are reported in Table~\ref{tab:bins_llama_human}.
Furthermore, to assess the reliability of the human evaluation, we measured inter-annotator agreement among annotators who evaluated the same subset.
In total, 543 atomic facts were evaluated, and the annotators agreed on 515 of them, corresponding to a 94.8\% raw agreement.
\section{Why Do Hallucinations Frequently Occur at the Last?}
\label{rq2}
This section investigates the underlying factors that contribute to the \textit{Hallucinate at the Last} phenomenon, exploring two main hypotheses. The first attributes the phenomenon to the \textit{inherent nature of summarization}: key information is typically concentrated at the beginning of the summary, with less important content appearing toward the end. The second posits that as LLMs generate longer outputs, they increasingly attend to previously generated tokens rather than the original input context, resulting in a \textit{shift in attention distribution}.

\subsection{Is it Intrinsic to Summarization?}

This hypothesis draws on the intrinsic structure of both human- and model-generated summaries.
In typical summarization domains such as news, scientific articles, and Wikipedia-style documents, key information is predominantly introduced early in the summary~\cite{10.5555/2969239.2969428, narayan-etal-2018-dont, cohan-etal-2018-discourse}, reflecting the \textbf{lead bias} observed in source texts, where the opening contains the most salient content~\cite{kim-etal-2019-abstractive, zhao-etal-2022-narrasum, ravaut-etal-2024-context}.
Building on this, our study investigates whether this structural bias is also reflected in decreasing factual consistency within generated outputs by comparing positional faithfulness scores between human-written and model-generated summaries.

\begin{figure}[!t]
\centering
\includegraphics[width=0.8\linewidth]{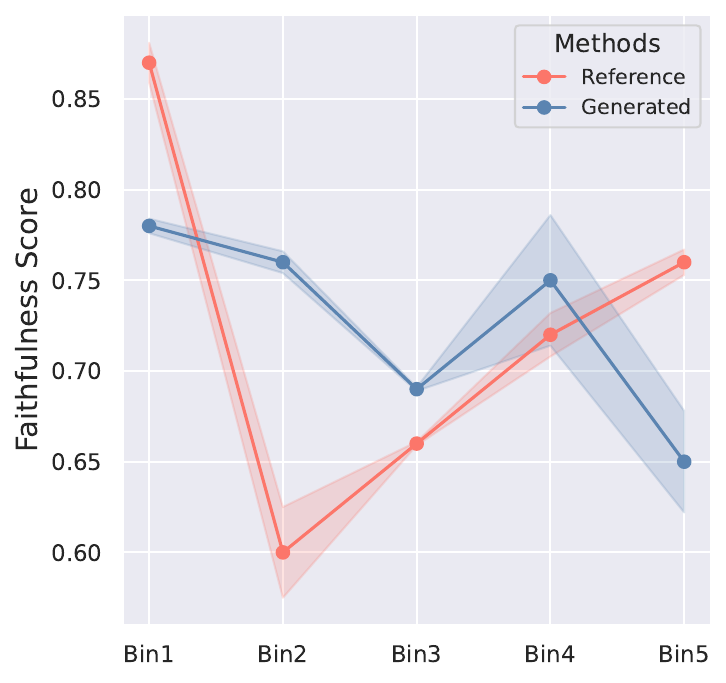}
\caption{Faithfulness scores of reference summaries and those generated by Llama on the CNNDM dataset.}
\label{fig5}
\vspace{-5mm}
\end{figure}

\paragraph{Experimental Setup}
We use human-written reference summaries from the CNNDM dataset, selecting examples with long reference summaries.
We compare these references to summaries generated by the Llama3.1-8B-Instruct model.

\paragraph{Results \& Analysis}

Figure~\ref{fig5} compares positional faithfulness between human-written and model-generated CNNDM summaries.
We observe that the faithfulness of the generated summaries declines steadily, falling below 0.65 in the final segment, whereas reference summaries dip around the middle but recover.
This suggests that \textit{Hallucination at the Last} \textbf{is not solely due to summarization’s inherent structure}.

\subsection{Is it Intrinsic to Attention?}
\label{ssec:intrinsic_attention}
Prior research on LLMs has demonstrated that attention weight distributions are closely correlated with the generation process, influencing output coherence~\cite{dong-etal-2021-fly, zhang2024tell}.
Recent studies on Large Vision-Language Models (LVLMs) further suggest that hallucinations frequently occur in the later parts of generated text~\cite{liu2024mitigating, lee2024toward, min-etal-2025-mitigating}.
This phenomenon has been attributed to a shift in attention: as text generation progresses, attention weights increasingly favor previously generated text tokens over image tokens.
Inspired by this observation, we investigate whether a similar trend exists in LLMs by analyzing how attention weights on generated tokens evolve as output length increases.

\paragraph{Experimental Setup}
Following \citet{hsieh-etal-2024-found}, we compute attention weights by averaging the decoder self-attention matrices across all layers and heads.
Unlike the prior study, however, we compute sentence-level attention by aligning token spans with sentence boundaries and averaging the token-level attention weights within each sentence.
We segment the sequence into 100-token chunks and focus our analysis on three positions: the first, middle, and final sentences.
Further details are provided in Appendix~\ref{sec:appx_attention_weight}.

\begin{figure}[!t]
\centering
\begin{subfigure}{\linewidth}
\includegraphics[width=0.95\linewidth]{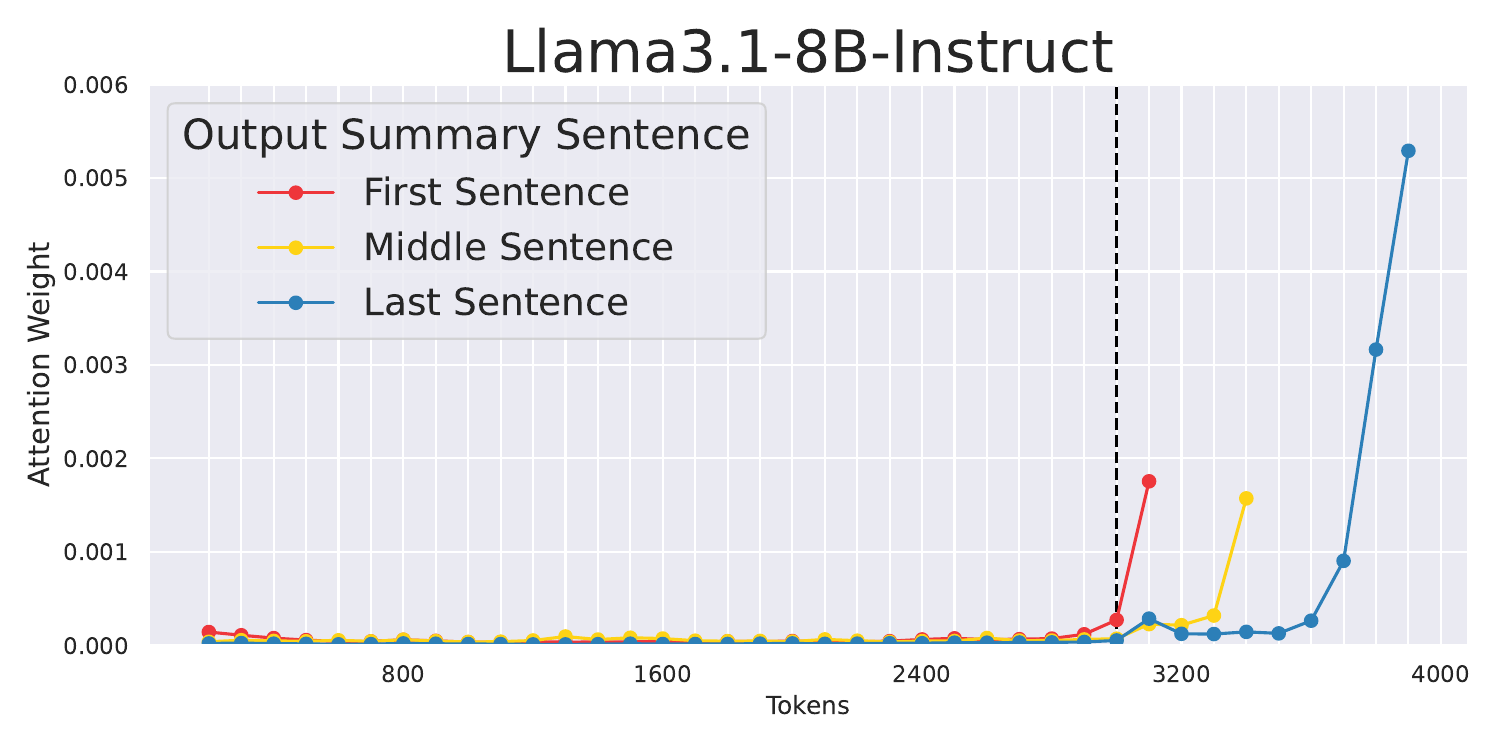}
\caption{Average attention weights of summaries generated by \textbf{Llama}.}
\label{fig:llama_attention}
\end{subfigure}
\begin{subfigure}{\linewidth}
\includegraphics[width=0.95\linewidth]{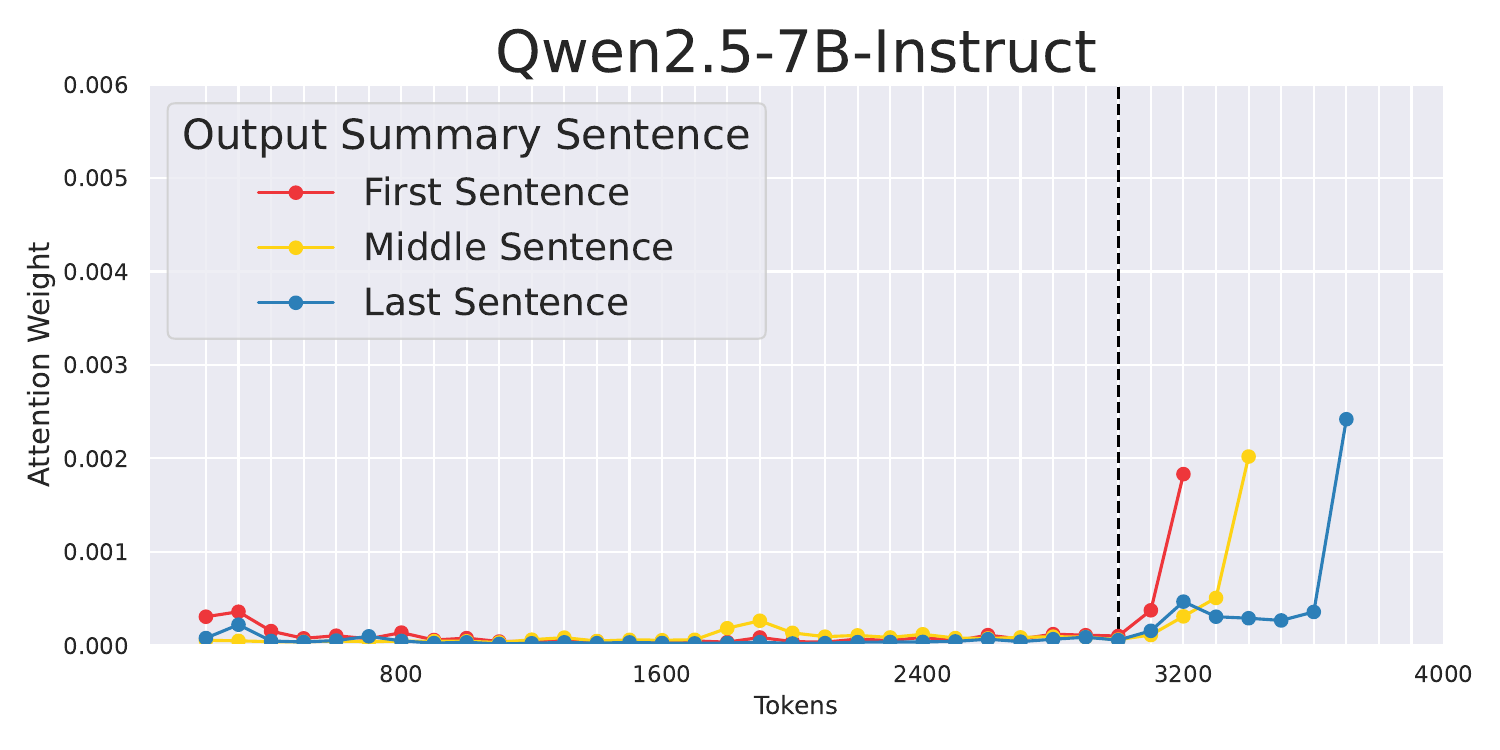}
\caption{Average attention weights of summaries generated by \textbf{Qwen} with sliding window attention.}
\label{fig:qwen_attention}
\end{subfigure}
\begin{subfigure}{\linewidth}
\includegraphics[width=0.95\linewidth]{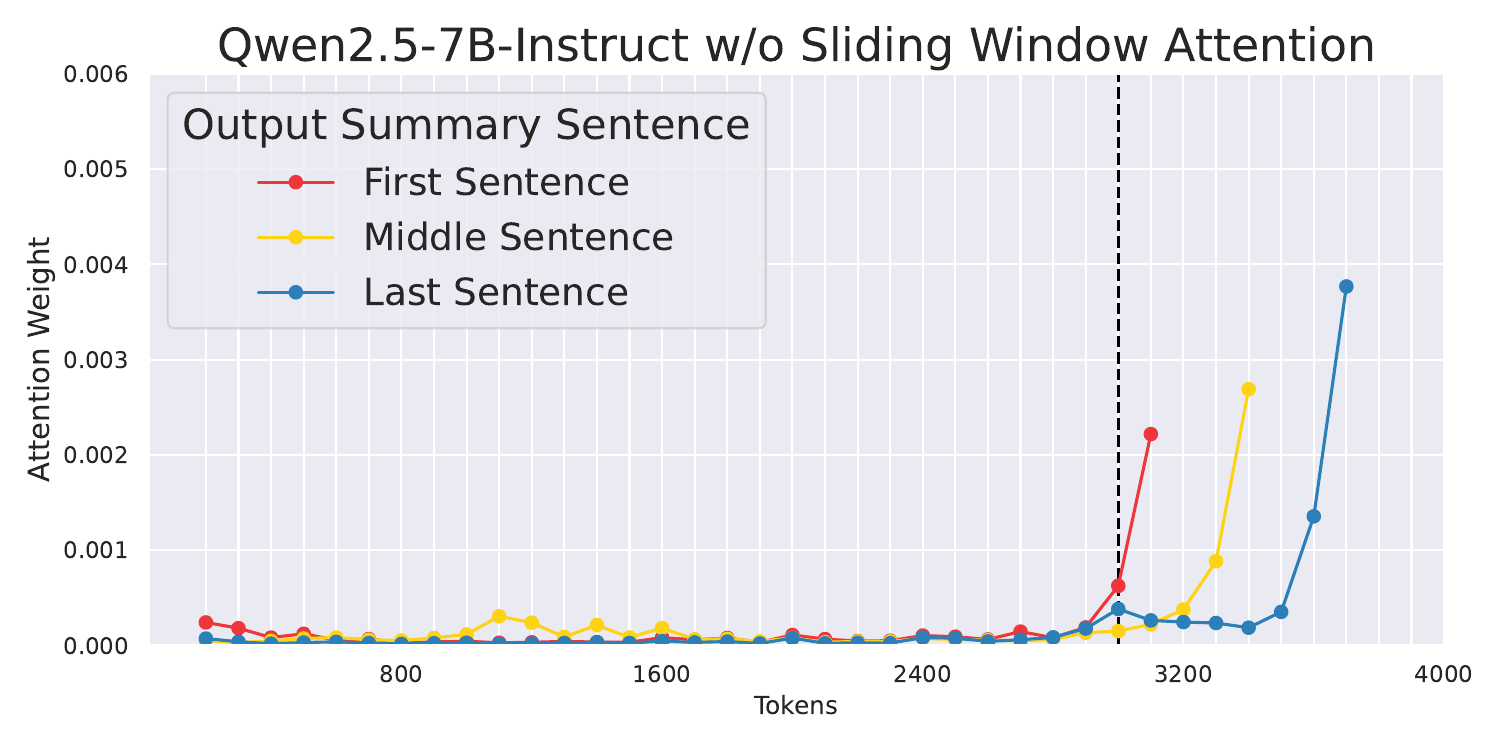}
\caption{Average attention weights of summaries generated by \textbf{Qwen} without sliding window attention.}
\label{fig:qwen_wo_attention}
\end{subfigure}
\caption{Average attention weights of two models exhibiting contrasting trends in Figure~\ref{fig2}. The dashed line separates the input context from the generated output.}
\label{fig6}
\vspace{-5mm}
\end{figure}

\paragraph{Results \& Analysis}
Figure~\ref{fig6} presents a visualization of average attention weights across sentences.
As shown in Figure~\ref{fig:llama_attention}, the Llama model, which exhibited the \textit{Hallucinate at the Last} pattern in Figure~\ref{fig2}, assigns nearly three times more attention to the final sentence of the generated summary compared to the first and middle sentences.
In contrast, the Qwen model (Figure~\ref{fig:qwen_attention}) assigns similar levels of attention to all three sentence positions.
These findings suggest that \textbf{increased attention to previously generated text correlates with a higher likelihood of hallucination}.



This consistent distribution in Qwen can be attributed to its sliding window attention mechanism~\cite{child2019generatinglongsequencessparse}, which maintains a balanced focus across the context and mitigates the downstream attention concentration that triggers hallucinations.
Table~\ref{tab:bins_qwen_minicheck_t5} compares Qwen with and without sliding window attention, where the sliding-window context length is fixed at 16,000 tokens.
The results clearly indicate that removing the sliding window attention mechanism increases the sensitivity to 3.75 for standard summaries and 3.25 for long summaries.
Furthermore, as shown in Figure~\ref{fig:qwen_wo_attention}, the attention level becomes higher than that observed with sliding window attention.
While autoregressive decoders generally allocate more attention to recent tokens, this trend is not uniform across models.
Taken together, these results demonstrate that sliding window attention not only enhances decoding efficiency but also \textbf{reduces attention concentration on later tokens as the generated sequence length increases}.

\begin{table}[!t]
    \centering
    \resizebox{\linewidth}{!}{
        \begin{tabular}{c|ccccc|c}
            \toprule
            \multirow{2}{*}{\textbf{\textsc{Models}}} & \multicolumn{5}{c|}{\textbf{\textsc{Generated Summary Bins}}} & \multirow{2}{*}{\textbf{\textsc{Sensitivity}}} \\
            & \textbf{\textsc{1}} & \textbf{\textsc{2}} & \textbf{\textsc{3}} & \textbf{\textsc{4}} & \textbf{\textsc{5}} & \\
            \midrule
            \multicolumn{7}{c}{\textbf{Standard Summary}} \\
            \midrule
            Qwen2.5-7B-Instruct & \textbf{0.89} & 0.88 & 0.85 & \underline{0.81} & 0.87 & -1.25 \\
            \textit{w/o sliding window attention}& \textbf{0.90} & 0.88 & 0.85 & \underline{0.83} & 0.84 & 2.5\textcolor{red}{(+3.75)} \\
            \midrule
            \multicolumn{7}{c}{\textbf{Long Summary}} \\
            \midrule
            Qwen2.5-7B-Instruct
            & \textbf{0.86} & 0.83 & 0.83 & \underline{0.82} & 0.84 & -0.50 \\
            \textit{w/o sliding window attention}& \textbf{0.86} & 0.84 & 0.83 & 0.82 & \underline{0.81} & 2.75\textcolor{red}{(+3.25)} \\
            \bottomrule
        \end{tabular}
    }
    \caption{Faithfulness scores and sensitivity for \textbf{standard} and \textbf{long} summaries generated by Qwen2.5-7B-Instruct on the Wikipedia dataset, with and without sliding window attention. Note that higher sensitivity indicates stronger hallucination effects toward the end of the output.}
\label{tab:bins_qwen_minicheck_t5}
\vspace{-5mm}
\end{table}

\section{How to Mitigate the \textit{Hallucinate at the Last}?}
\label{rq3}

In the preceding section, we demonstrated that the \textit{Hallucinate at the Last} phenomenon could be partially mitigated through the use of sliding window attention.
In the following analysis, we investigate strategies to address this issue in models that are not trained with sliding window attention.
To this end, we apply four methods that may mitigate the phenomenon.

\paragraph{Experimental Setup}
In this experiment, we generate Wikipedia summaries using four methods, taking Llama3.1-8B-Instruct as the baseline.
For comparison, we evaluate the following four methods: (See Appendix~\ref{sec:appx_mitigation} for more experimental details.)
\begin{itemize}[leftmargin=*,itemsep=0.1em]
    \item \textbf{\textsc{BooookScore}}~\cite{chang2024booookscore} segments the input context into chunks, generates summaries for each chunk individually, and then merges the partial summaries.
    \item \textbf{\textsc{MInference}}~\cite{jiang2024minference} employs sparse attention mechanisms to efficiently process long input sequences.
    \item \textbf{\textsc{LongWriter-Llama3.1-8B}}~\cite{bai2025longwriter} is a model fine-tuned on a long-output dataset and further enhanced using DPO~\cite{rafailov2023direct}.
    \item \textbf{\textsc{AdaCAD}}~\cite{wang-etal-2025-adacad} enhances factual consistency during generation by context-aware decoding~\cite{shi-etal-2024-trusting}.
\end{itemize}



\begin{table}[!b]
    \centering
    \resizebox{\linewidth}{!}{
        \begin{tabular}{l|ccccc|c}
            \toprule
            \multirow{2}{*}{\textbf{\textsc{Methods}}} & \multicolumn{5}{c|}{\textbf{\textsc{Generated Summary Bins}}} & \multirow{2}{*}{\textbf{\textsc{Sensitivity}}} \\
             & \textbf{\textsc{1}} & \textbf{\textsc{2}} & \textbf{\textsc{3}} & \textbf{\textsc{4}} & \textbf{\textsc{5}} & \\
            \midrule
            Llama3.1-8B & 0.74 & \textbf{0.75} & \textbf{0.75} & 0.73 & \underline{0.64} & 10.3 \\
            + \textsc{BooookScore} & 0.73 & 0.77 & 0.78 & \textbf{0.80} & 0.77 & 0.0 \\
            + \textsc{MInference} & 0.77 & \textbf{0.78} & \textbf{0.78} & 0.71 & \underline{0.69} & 7.0 \\
            + \textsc{LongWriter} & \textbf{0.85} & 0.77 & 0.81 & \underline{0.63} & 0.70 & 6.5 \\
            + \textsc{AdaCAD} & 0.79 & \textbf{0.82} & 0.77 & 0.81 & \underline{0.64} & 15.8 \\
            \bottomrule
        \end{tabular}
    }
    \caption{Faithfulness scores and sensitivity for \textbf{long} summaries generated by different mitigation methods on the Wikipedia dataset.}
\label{tab:bins_mitigation}
\vspace{-5mm}
\end{table}

\begin{figure}[!t]
\centering
\includegraphics[width=0.96\linewidth]{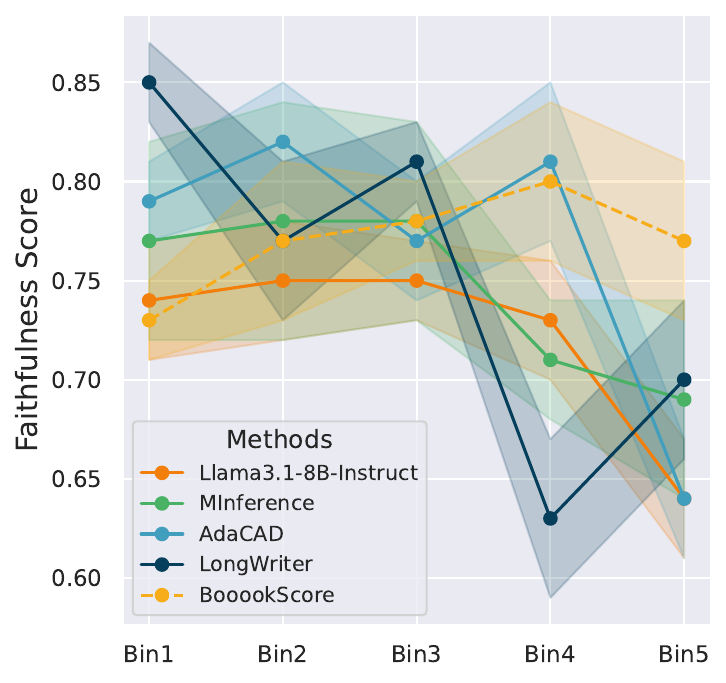}
\caption{Faithfulness scores for \textbf{long} summaries generated by different mitigation methods on the Wikipedia dataset across increasing output lengths.}
\label{fig7}
\vspace{-5mm}
\end{figure}

\paragraph{Results \& Analysis}
We report the faithfulness scores across output bins and the corresponding sensitivity values in Table~\ref{tab:bins_mitigation}, and present the bin-wise faithfulness trends for each method in Figure~\ref{fig7}.
As shown in Table~\ref{tab:bins_mitigation}, \textsc{BooookScore} achieves the lowest sensitivity (0.0), making it the closest to zero among the four methods.
Moreover, as illustrated in Figure~\ref{fig7}, \textsc{BooookScore} maintains the highest level of faithfulness, particularly around the final bin, and notably avoids the sharp decline in faithfulness observed in other methods at the end of the summary.
These results suggest that \textbf{generating summaries independently for each chunk and subsequently merging them can be an effective strategy} for mitigating the \textit{Hallucinate at the Last} phenomenon.
In addition, we report results for other evaluation dimensions on the summaries generated using \textsc{BooookScore} in Table~\ref{tab:tab_boookscore}.
The results show that \textbf{\textsc{BooookScore} not only mitigates the \textit{Hallucinate at the Last} phenomenon, but also improves the overall quality of the generated summaries.}

Nevertheless, while \textsc{BooookScore} provides a practical mitigation for models without sliding window attention, it does not fundamentally resolve the issue, as it relies on chunking and merging rather than direct long-context generation.
This underscores the need for future methods that can maintain faithfulness when generating long text directly from long contexts.
\section{Related Work}

The problem of hallucination in LLMs has been a significant area of research.
Prior work has largely focused on developing methods for detecting factual inconsistencies in generated text~\cite{chuang-etal-2024-lookback, hu2024refcheckerreferencebasedfinegrainedhallucination, kim2024fables, zhong-litman-2025-discourse} and proposing strategies to mitigate their occurrence, often through improved training~\cite{zhang-etal-2023-merging, wan-etal-2023-histalign}, decoding~\cite{shi-etal-2024-trusting, wang-etal-2025-adacad}, or prompting~\cite{zhou-etal-2023-context}.
In addition, \citet{wan-etal-2025-positional} analyzed positional hallucinations in long-context summarization.
While these approaches reduce overall hallucination rates, most exisiting studies do not examine how hallucinations are distributed within the generated sequence, especially in long-form outputs.

Positional bias in generation errors has also been observed in related domains. In particular, studies on Large Vision–Language Models (LVLMs) report that hallucinations tend to increase toward the end of generated image descriptions~\cite{liu2024mitigating, lee2024toward, min-etal-2025-mitigating}.

Recently, several studies have begun to explore long-output generation in LLMs~\cite{wu2025longgenbench, ye2025longprocbenchmarkinglongcontextlanguage}.
However, these works focus on procedural generation tasks that are not grounded in contextual input, limiting their applicability to real-world summarization scenarios.
A notable study on multi-document summarization aligns with our findings~\cite{belem-etal-2025-single, zhao2025doesresponselengthaffect}.
Nevertheless, it does not address positional context-aware faithfulness over extended summary lengths.

Our work presents the first dedicated study of where hallucinations occur in LLM-based long document summarization, moving beyond detection to uncover positional patterns, underlying causes, and mitigation strategies.
\begin{table}[!t]
    \centering
    \resizebox{\linewidth}{!}{
        \begin{tabular}{l|c|c|c}
            \toprule
            \textbf{\textsc{Methods}} & \textbf{\textsc{Faithfulness(\%)}} & \textbf{\textsc{Completeness(\%)}} & \textbf{\textsc{Conciseness(\%)}} \\
            \midrule
            Llama3.1-8B & 90.33 & 100.0 & 41.35 \\
            \textsc{BooookScore} & 94.33 & 100.0 & 47.83 \\
            \bottomrule
        \end{tabular}
    }
    \caption{Overall faithfulness, completeness and conciseness scores of \textbf{long} summaries on the Wikipedia dataset, evaluated using FineSurE.}
\label{tab:tab_boookscore}
\vspace{-5mm}
\end{table}

\section{Conclusion}
We identified and characterized the \textit{Hallucinate at the Last} phenomenon in LLM-based long response generation, specifically in long document summarization.
Our findings show that hallucinations disproportionately increase towards the end of long outputs, a bias amplified in longer summaries.
We investigated the contributing factors and explored targeted mitigation strategies.
Our work highlighted the importance of the output's faithfulness and motivates future research into positionally-aware generation techniques.
\section*{Limitations}

In this study, we explored four domains, primarily because the corresponding datasets provided input contexts of the desired length.
However, we were not able to investigate domains such as books, dialogues, movie scripts, or meeting transcripts.
Books were excluded due to their excessive length, while the other domains lacked datasets with suitable input lengths or sufficient sample sizes.

Despite these limitations, we believe this work offers an important foundation for future research in the relatively underexplored domain of long-form output generation, particularly in summarization.
\section*{Ethics Statement}
 
This study leverages publicly available datasets, including Wikipedia, Arxiv, Pubmed, Govreport, and CNNDM, to analyze long-form text generation in LLMs.
All experiments were conducted in a consistent and reproducible manner across models and datasets, without any manipulation or omission of data or results.
We investigate the phenomenon of \textit{Hallucinate at the Last}, a tendency observed in long context-aware summarization models to generate hallucinated content toward the final portion of the summary.
By drawing attention to this issue, our study contributes to ongoing efforts to enhance the reliability and factual consistency of LLM-generated summaries.
\section*{Acknowledgement}
This work was supported by the Institute of Information \& Communications Technology Planning \& Evaluation (IITP) grant funded by the Korea government (MSIT) [RS-2021-II211341, Artificial Intelligence Graduate School Program (Chung-Ang University)].

\bibliography{main}
\cleardoublepage
\appendix

\section{Attention Weight Calculation}
\label{sec:appx_attention_weight}
This section provides a more detailed and formal explanation of our attention weight computation method introduced in Section~\ref{rq2}.

Let the full sequence, consisting of the input context and the generated output summary, be defined as:
\begin{equation}
x_{full} = x_{prompt} + x_{output} = [x_1, x_2, ..., x_L]
\end{equation}
where $L$ is the number of tokens.

We partition $x_{full}$ into $K$ non-overlapping blocks of 100 tokens each, such that each block is represented as:
\begin{equation}
x^{k} = \{x_i^k\}_{i=1}^{100}
\end{equation}
for $k\in\{1, 2, ..., K\}$.

Let the output summary contain three target sentences: the \textbf{first}, \textbf{middle}, and \textbf{last} sentence, denoted respectively as:
\begin{equation}
\begin{aligned}
x^{first} &= \{x^{first}_j\}_{j=1}^{T_1}\\
x^{middle} &= \{x^{second}_j\}_{j=1}^{T_2}\\
x^{last} &= \{x^{third}_j\}_{j=1}^{T_3}
\end{aligned}
\end{equation}
where $T_1$, $T_2$, $T_3$ are the respective sentence lengths.

We define the average attention weight from block $x^k$ to each of the three sentences as follows:
{\small
\begin{equation}
\begin{gathered}
\textrm{\texttt{Attn}}_{first}(k) = \frac{1}{100\cdot{T_1}}\sum_{i=1}^{100}\sum_{j=1}^{T_1}\textrm{\texttt{attn}}(x_i^k \to x_j^{first}) \\
\textrm{\texttt{Attn}}_{middle}(k) = \frac{1}{100\cdot{T_2}}\sum_{i=1}^{100}\sum_{j=1}^{T_2}\textrm{\texttt{attn}}(x_i^k \to x_j^{middle}) \\
\textrm{\texttt{Attn}}_{last}(k) = \frac{1}{100\cdot{T_3}}\sum_{i=1}^{100}\sum_{j=1}^{T_3}\textrm{\texttt{attn}}(x_i^k \to x_j^{last})
\end{gathered}
\end{equation}
}

\section{Details for Atomic Fact Generation}
\label{sec:appx_atomic_fact}
This section details our experimental details and human evaluation results from Section~\ref{sec:generating_summaries}.

\paragraph{Prompt}
The prompt template employed for atomic fact generation is shown in Table~\ref{tab:tab_prompt_af}.

\paragraph{Human Evaluation}
Furthermore, we assess the quality of the atomic facts generated by the LLM through human evaluation, as reported in Table~\ref{tab:tab_af_valiation}, focusing on atomic facts extracted from long summaries generated by Llama.
For each context length, we randomly sampled 20 summaries, and the authors manually conducted the evaluation.
Each atomic fact was independently annotated by the authors as either true or false, depending on whether it was supported by the corresponding summary.
In total, 1,569 atomic facts were evaluated, and the two annotators agreed on 1,541 of them, corresponding to a 98.2\% raw agreement.
Because the task involves binary labels with a skewed class distribution, chance-corrected agreement coefficients are known to become unstable or difficult to interpret.
We therefore report raw agreement as a clearer and more reliable indicator of annotation consistency in this setting.
\begin{table}[!t]
    \centering
    \resizebox{0.6\linewidth}{!}{
        \begin{tabular}{c|c}
            \toprule
            \textbf{\textsc{Context Length}} & \textbf{\textsc{Validation(\%)}} \\
            \midrule
            4K & 95.92 \\
            5K & 99.01 \\
            6K & 96.18 \\
            7K & 99.12 \\
            8K & 99.64 \\
            \bottomrule
        \end{tabular}
    }
    \caption{Human evaluation of the quality of atomic facts.}
\label{tab:tab_af_valiation}
\vspace{-3mm}
\end{table}

\section{Details for Model Usage}
\label{sec:appx_evaluation_metric}
In this section, we present the experimental details for the model usage in Section~\ref{rq1}.

\paragraph{Details for Evaluation Method}
For atomic fact decomposition, we utilize the Llama3.1-8B-Instruct model with vLLM~\cite{kwon2023efficient}\footnote{\url{https://github.com/vllm-project/vllm}}.
To further illustrate the evaluation process, we present in Table~\ref{tab:example} a concrete example that demonstrates how a given sentence is decomposed into its constituent atomic facts and subsequently assessed for faithfulness.
This example clarifies the step-by-step procedure and provides transparency regarding the alignment between the generated atomic facts and the underlying reference information.

\begin{table}[!t]
    \centering
    \resizebox{\linewidth}{!}{
        \begin{tabular}{l|c}
            \toprule
            \multicolumn{2}{c}{\textbf{First Sentence}} \\
            \midrule
            \multicolumn{2}{p{12cm}}{
                Sara Sorribes Tormo is a talented Spanish professional tennis player born on October 8, 1996, in Castellón de la Plana, Spain.
            } \\
            \midrule
            \textbf{\textsc{Atomic Facts}} & \textbf{\textsc{Score}} \\
            \midrule
            Sara Sorribes Tormo is a Spanish professional tennis player. & 0.92 \\
            Sara Sorribes Tormo was born in Spain. & 0.90 \\
            Sara Sorribes Tormo was born on October 8. & 0.87 \\
            Sara Sorribes Tormo was born in 1996. & 0.96 \\
            Sara Sorribes Tormo was born in Castellón de la Plana. & 0.93 \\
            \midrule
            \textbf{Avg. Score} & \textbf{0.92} \\
            \midrule
            \multicolumn{2}{c}{\textbf{Last Sentence}} \\
            \midrule
            \multicolumn{2}{p{12cm}}{
                As Sara navigates her career, her unique blend of skill, perseverance, and spirit positions her as a promising athlete, inspiring both current and aspiring players.
            } \\
            \midrule
            \textbf{\textsc{Atomic Facts}} & \textbf{\textsc{Score}} \\
            \midrule
            Sara navigates her career. & 0.87 \\
            Sara’s career is that of an athlete. & 0.91 \\
            Sara has a unique blend of skill. & 0.14 \\
            Sara has perseverance. & 0.81 \\
            Sara is a promising athlete. & 0.30 \\
            Sara inspires current players. & 0.11 \\
            Sara inspires aspiring players. & 0.15 \\
            \midrule
            \textbf{Avg. Score} & \textbf{0.47} \\
            \bottomrule
        \end{tabular}
    }
    \caption{Example of the evaluation procedure with the MiniCheck-Flan-T5-Large Model.}
\label{tab:example}
\vspace{-5mm}
\end{table}

\paragraph{Details for LLMs Usage}
We utilized publicly available instruction-tuned models, including Llama3.1-8B-Instruct\footnote{\url{https://huggingface.co/meta-llama/Llama-3.1-8B-Instruct}}, Gemma3-12B-Instruct\footnote{\url{https://huggingface.co/google/gemma-3-12b-it}}, and Qwen2.5-7B-Instruct\footnote{\url{https://huggingface.co/Qwen/Qwen2.5-7B-Instruct}} from HuggingFace.
For the GPT-4o mini model, we used \texttt{gpt-4o-mini-2024-07-18}.
All summaries were generated using greedy decoding with \texttt{float16} precision.

\paragraph{Details for Bins}
Throughout all experiments, we used five bins.
The bins were computed using the \texttt{cut} function in the pandas package, and the scores of the atomic facts were assigned to bins using open interval boundaries, meaning that the boundary values were not included in the bins.

To ensure that our findings are not the result of the specific binning choice, we additionally report results using a finer-grained setting with ten bins in Table~\ref{tab:bins_ten}.
The same trend persists even with this increased resolution, confirming that the \textit{Hallucinate at the Last} phenomenon is robust to the choice of bin granularity.

\begin{table*}[!h]
    \centering
    \small
    \resizebox{\linewidth}{!}{
        \begin{tabular}{l|cccccccccc|c}
            \toprule
            \multirow{2}{*}{\textbf{\textsc{Models}}} & \multicolumn{10}{c|}{\textbf{\textsc{Generated Summary Bins}}} & \multirow{2}{*}{\textbf{\textsc{Sensitivity}}} \\
             & \textbf{\textsc{1}} & \textbf{\textsc{2}} & \textbf{\textsc{3}} & \textbf{\textsc{4}} & \textbf{\textsc{5}} & \textbf{\textsc{6}} & \textbf{\textsc{7}} & \textbf{\textsc{8}} & \textbf{\textsc{9}} & \textbf{\textsc{10}} & \\
            \midrule
            Llama3.1-8B & 0.80 & 0.79 & \textbf{0.83} & 0.77 & 0.71 & 0.74 & 0.72 & 0.83 & 0.76 & \underline{0.70} & 7.22 \\
            Qwen2.5-7B & 0.85 & 0.85 & \underline{0.80} & 0.82 & 0.85 & 0.82 & \textbf{0.89} & 0.81 & 0.89 & 0.86 & -1.78 \\
            \bottomrule
        \end{tabular}
    }
    \caption{Faithfulness scores and sensitivity for \textbf{long} summaries generated by Llama and Qwen.}
\label{tab:bins_ten}
\vspace{-5mm}
\end{table*}

\section{More Results with Different Methods}
\label{sec:appx_different_methods}
We adopt a diverse set of methods to validate the robustness of our faithfulness evaluation framework.
In this section, we present experimental results for faithfulness assessment models beyond MiniCheck, incorporating approaches based on NLI, ROUGE-L, and LLMs.
Specifically, we substitute the scoring model $\mathcal{M}$ in Equation~\ref{eq:eq2} of Section~\ref{rq1} with these alternative methods.

\paragraph{Minicheck Based}
We utilize MiniCheck-Flan-T5-Large Model\footnote{\url{https://huggingface.co/lytang/MiniCheck-Flan-T5-Large}} for the main evaluation method.
We report the faithfulness scores and sensitivity for \textbf{standard} summaries generated by different models across output bins on the Wikipedia dataset in Table~\ref{tab:bins_overall_models_short_minicheck_t5}, and those of \textbf{long} summaries in Table~\ref{tab:bins_overall_models_long_minicheck_t5}.

\paragraph{NLI Based}
For the NLI model, we use the state-of-the-art hallucination evaluation model~\cite{hhem-2.1-open}\footnote{\url{https://huggingface.co/vectara/hallucination_evaluation_model}}.

We report the faithfulness scores and sensitivity assessed by NLI model for \textbf{standard} summaries generated by different models across output bins on the Wikipedia dataset in Table~\ref{tab:bins_overall_models_short}, and those of \textbf{long} summaries in Table~\ref{tab:bins_overall_models_long}.
As observed in the experimental results, consistent with the main results, the Llama and GPT models consistently exhibit high sensitivity, whereas the Qwen model occasionally demonstrates sensitivity values even below zero.

\paragraph{ROUGE Based}
We report the faithfulness scores and sensitivity assessed by ROUGE-L~\cite{lin-2004-rouge} for \textbf{standard} summaries generated by different models across output bins on the Wikipedia dataset in Table~\ref{tab:bins_overall_models_short_rougel}, and those of \textbf{long} summaries in Table~\ref{tab:bins_overall_models_long_rougel}.

\paragraph{LLM Based}
We present the error rates assessed by FineSurE framework for \textbf{standard} summaries generated by different models across output bins on the Wikipedia dataset in Table~\ref{tab:bins_overall_models_short_finesure}, and those of \textbf{long} summaries in Table~\ref{tab:bins_overall_models_long_finesure}.

As observed in the experimental results, consistent with the main results, the Llama and GPT models consistently exhibit the highest error rates in the final bin.
In contrast, the Qwen model does not yield the highest error rates in the final bin for both the cases of standard and long summaries.





\section{More Results on Varying Datasets}
\label{sec:appx_varying_datasets}

\paragraph{Minicheck Based}
We report the faithfulness scores and sensitivity assessed by MiniCheck-Flan-T5-Large model of \textbf{standard} summaries generated by the Llama3.1-8B-Instruct model across output bins in multiple domains in Table~\ref{tab:bins_overall_datasets_short_minicheck_t5}, and those of \textbf{long} summaries in Table~\ref{tab:bins_overall_datasets_long_minicheck_t5}.

The experimental results reveal that in the case of standard summaries, sensitivities below zero are frequently observed; notably, in the Pubmed dataset, more than half of the values fall below zero.
In contrast, such negative sensitivities are rarely observed in long summaries, with the Pubmed dataset in particular showing no sensitivities below zero, unlike the standard summary setting.

\paragraph{NLI Based}
We report the faithfulness scores and sensitivity assessed by NLI model for \textbf{standard} summaries generated by the Llama3.1-8B-Instruct model across output bins in multiple domains in Table~\ref{tab:bins_overall_datasets_short}, and those of \textbf{long} summaries in Table~\ref{tab:bins_overall_datasets_long}.

As demonstrated by the experimental results, consistent with the main results, none of the summaries generated by the Llama3.1-8B-Instruct model across various domains exhibited a sensitivity value below zero.

where $\textrm{\texttt{attn}}(x_i \to x_j)$ denotes the attention weight from token $x_i$ to token $x_j$ when generating the output.

These measures quantify the extent to which each 100-token block in the full sequence contributes to the generation of the respective sentences in the output summary.

\section{Details for Mitigation Methods}
\label{sec:appx_mitigation}
In this section, we provide the experimental details of the mitigation methods described in Section~\ref{rq3}.

For the \textsc{BooookScore} method, the input is divided into 2048-token chunks, summaries are generated for each chunk using the Llama3.1-8B-Instruct model, and the partial summaries are then hierarchically merged into a final summary.
For \textsc{LongWriter}, we employed the \textsc{LongWriter-Llama3.1-8B} model\footnote{\url{https://huggingface.co/THUDM/LongWriter-llama3.1-8b}} with greedy decoding and \texttt{bfloat16} precision.

\section{Details for Generating Summaries}
\label{sec:appx_generating_summaries}
Building upon Section~\ref{sec:generating_summaries}, this section details our approach to generating long summaries and presents an analysis of the generated outputs.

\paragraph{Prompt}
Standard outputs were standardized to a length of 100 to 200 words, while long outputs were approximated by converting 30\% of the total context token length into a corresponding word range.
The specific word ranges used for prompting long-summary generation are provided in Table~\ref{tab:tab_words}, and the prompt template employed for summary generation is shown in Table~\ref{tab:tab_prompt}.

\begin{table}[!h]
    \centering
    \resizebox{0.7\linewidth}{!}{
        \begin{tabular}{c|c}
            \toprule
            \textbf{\textsc{Context Length}} & \textbf{\textsc{Words Range}} \\
            \midrule
            4K & 800 to 1000 \\
            5K & 1000 to 1250 \\
            6K & 1200 to 1500 \\
            7K & 1400 to 1750 \\
            8K & 1600 to 2000 \\
            \bottomrule
        \end{tabular}
    }
    \caption{Words range used in the prompt for long output generation.}
\label{tab:tab_words}
\end{table}

\paragraph{Generated Summary Length}
Our experiments revealed that controlling the output length of LLMs remains a significant challenge.
This difficulty becomes more pronounced as the input context length increases and as the target word range specified in the prompt becomes larger.
In particular, with longer contexts, LLMs often produce excessively long summaries containing repeated sentences or even entire paragraphs.
To ensure fair and reliable evaluation, we excluded such outputs and considered only summaries with comparable lengths in our analysis.
Specifically, we compared summaries within each sentence-count range and selected those with the highest sentence counts as length-matched summaries.
For example, in our setting, the length-matched summaries correspond to 28 sentences for the 4K-context case and 40 sentences for the 8K-context case.

Table~\ref{tab:tab_models_words} presents the average word counts of both standard and long summaries generated by each model.
As the results indicate, \textsc{LongWriter-Llama3.1-8B} frequently produces considerably longer outputs, while Qwen2.5-7B-Instruct tends to generate relatively shorter summaries.

\begin{table}[!t]
    \centering
    \resizebox{0.9\linewidth}{!}{
        \begin{tabular}{l|c|c}
            \toprule
            \multirow{2}{*}{\textbf{\textsc{Models}}} & \multicolumn{2}{c}{\textbf{\textsc{Avg. Words}}} \\
            & standard & long \\
            \midrule
            Llama3.1-8B-Instruct & 279$\pm{24}$ & 834$\pm{399}$ \\
            Qwen2.5-7B-Instruct & 153$\pm{39}$ & 751$\pm{204}$ \\
            Mistral-7B-Instruct-v0.3 & 285$\pm{292}$ & 794$\pm{390}$ \\
            GPT-4o mini & 172$\pm{8}$ & 925$\pm{132}$ \\
            LongWriter-Llama3.1-8B & 398$\pm{171}$ & 3291$\pm{979}$ \\
            \bottomrule
        \end{tabular}
    }
    \caption{The average word counts of both \textbf{standard} and \textbf{long} summaries generated by each model on the Wikipedia dataset.}
\label{tab:tab_models_words}
\end{table}

\section{Use of Large Language Model}
This paper primarily utilized the AI assistant (ChatGPT-5.1) for grammar correction and refinement of clarity.
It was not employed for content generation or research ideation.

\begin{table*}[h]
{\scriptsize
\centering
\setlength{\tabcolsep}{5pt}
\renewcommand{\arraystretch}{1.2}

\resizebox{\linewidth }{!}{

\begin{tabular}{p{16cm}}
\toprule
\textbf{Input Prompt} \\
\midrule
Write an accurate and engaging summary for the given text in range of \{\{words\_range\}\} words using only the provided passage (might be irrelevant).

Use an unbiased and journalistic tone.

Text: {\{\{Text\}\}}
\\
\bottomrule
\end{tabular}
}
\caption{Prompt template used to generate summaries from the original document in Section~\ref{sec:generating_summaries}.}
\label{tab:tab_prompt}
}
\end{table*}
\begin{table*}[h]
{\scriptsize
\centering
\setlength{\tabcolsep}{5pt}
\renewcommand{\arraystretch}{1.2}

\resizebox{\linewidth }{!}{

\begin{tabular}{p{16cm}}
\toprule
\textbf{Input Prompt} \\
\midrule
Decompose the text into atomic facts.

{\{\{Text\}\}}
\\
\bottomrule
\end{tabular}
}
\caption{Prompt template used to generate atomic facts from the summary in Section~\ref{sec:generating_summaries}.}
\label{tab:tab_prompt_af}
}
\end{table*}

\clearpage

\begin{table*}[h]
    \centering
    \resizebox{0.7\linewidth}{!}{
        \begin{tabular}{c|c|ccccc|c}
            \toprule
            \multirow{2}{*}{\textbf{\textsc{Models}}} & \textbf{\textsc{Context}} & \multicolumn{5}{c|}{\textbf{\textsc{Generated Summary Bins}}} & \multirow{2}{*}{\textbf{\textsc{Sensitivity}}} \\
            & \textbf{\textsc{Length}} & \textbf{\textsc{1}} & \textbf{\textsc{2}} & \textbf{\textsc{3}} & \textbf{\textsc{4}} & \textbf{\textsc{5}} & \\
            \midrule
            \multirow{8}{*}{Llama3.1-8B-Instruct} & 1K & 0.81 & \underline{0.79} & \textbf{0.85} & 0.82 & \underline{0.79} & 2.75 \\
            & 2K & \textbf{0.84} & 0.79 & 0.81 & 0.81 & \underline{0.77} & 4.25 \\
            & 3K & \textbf{0.87} & 0.85 & 0.84 & 0.83 & \underline{0.79} & 5.75 \\
            & 4K & \textbf{0.91} & 0.90 & 0.88 & \underline{0.86} & 0.88 & 0.75 \\
            & 5K & 0.85 & 0.85 & \textbf{0.86} & \underline{0.83} & \underline{0.83} & 1.75 \\
            & 6K & \textbf{0.89} & 0.85 & 0.85 & 0.83 & \underline{0.81} & 4.5 \\
\bottomrule
        \end{tabular}
    }
    \caption{Comparison of MiniCheck-Flan-T5-Large based faithfulness scores and Sensitivity for \textbf{standard} summaries generated by different models across output bins on the Wikipedia dataset, with context lengths ranging from 1K to 8K. The highest faithfulness score in each bin is marked in \textbf{bold}, while the lowest is \underline{underlined}.}
\label{tab:bins_overall_models_short_minicheck_t5}
\end{table*}
\begin{table*}[h]
    \centering
    \resizebox{0.7\linewidth}{!}{
        \begin{tabular}{c|c|ccccc|c}
            \toprule
            \multirow{2}{*}{\textbf{\textsc{Models}}} & \textbf{\textsc{Context}} & \multicolumn{5}{c|}{\textbf{\textsc{Generated Summary Bins}}} & \multirow{2}{*}{\textbf{\textsc{Sensitivity}}} \\
            & \textbf{\textsc{Length}} & \textbf{\textsc{1}} & \textbf{\textsc{2}} & \textbf{\textsc{3}} & \textbf{\textsc{4}} & \textbf{\textsc{5}} & \\
            \midrule
            \multirow{8}{*}{Llama3.1-8B-Instruct} & 1K & 0.78 & \textbf{0.82} & \underline{0.75} & \underline{0.75} & \underline{0.75} & 2.5 \\
            & 2K & \textbf{0.82} & 0.79 & 0.75 & 0.76 & \underline{0.73} & 5.0 \\
            & 3K & 0.77 & \textbf{0.81} & 0.80 & \textbf{0.81} & \underline{0.71} & 8.75 \\
            & 4K & 0.79 & \textbf{0.80} & \underline{0.73} & 0.77 & 0.74 & 3.25 \\
            & 5K & 0.80 & 0.83 & 0.83 & \textbf{0.84} & \underline{0.79} & 3.5 \\
            & 6K & 0.80 & 0.82 & 0.77 & \textbf{0.83} & \underline{0.72} & 8.5 \\
\bottomrule
        \end{tabular}
    }
    \caption{Comparison of MiniCheck-Flan-T5-Large based faithfulness scores and Sensitivity for \textbf{long} summaries generated by different models across output bins on the Wikipedia dataset, with context lengths ranging from 1K to 8K. The highest faithfulness score in each bin is marked in \textbf{bold}, while the lowest is \underline{underlined}.}
\label{tab:bins_overall_models_long_minicheck_t5}
\end{table*}
\begin{table*}[h]
    \centering
    \resizebox{0.7\linewidth}{!}{
        \begin{tabular}{c|c|ccccc|c}
            \toprule
            \multirow{2}{*}{\textbf{\textsc{Models}}} & \textbf{\textsc{Context}} & \multicolumn{5}{c|}{\textbf{\textsc{Generated Summary Bins}}} & \multirow{2}{*}{\textbf{\textsc{Sensitivity}}} \\
            & \textbf{\textsc{Length}} & \textbf{\textsc{1}} & \textbf{\textsc{2}} & \textbf{\textsc{3}} & \textbf{\textsc{4}} & \textbf{\textsc{5}} & \\
            \midrule
            \multirow{8}{*}{Llama3.1-8B-Instruct} & 1K & 0.73 & 0.73 & \textbf{0.75} & 0.74 & \underline{0.71} & 3.0 \\
            & 2K & \textbf{0.77} & 0.76 & 0.75 & \underline{0.73} & \underline{0.73} & 2.25 \\
            & 3K & 0.79 & \textbf{0.80} & 0.76 & 0.76 & \underline{0.69} & 8.75 \\
            & 4K & 0.81 & \textbf{0.82} & 0.81 & 0.79 & \underline{0.76} & 4.75 \\
            & 5K & \textbf{0.80} & 0.79 & 0.79 & 0.76 & \underline{0.75} & 3.5 \\
            & 6K & \textbf{0.81} & 0.80 & 0.80 & 0.76 & \underline{0.71} & 8.25 \\
\bottomrule
        \end{tabular}
    }
    \caption{Comparison of NLI model based faithfulness scores and Sensitivity for \textbf{standard} summaries generated by different models across output bins on the Wikipedia dataset, with context lengths ranging from 1K to 8K. The highest faithfulness score in each bin is marked in \textbf{bold}, while the lowest is \underline{underlined}.}
\label{tab:bins_overall_models_short}
\end{table*}
\begin{table*}[h]
    \centering
    \resizebox{0.7\linewidth}{!}{
        \begin{tabular}{c|c|ccccc|c}
            \toprule
            \multirow{2}{*}{\textbf{\textsc{Models}}} & \textbf{\textsc{Context}} & \multicolumn{5}{c|}{\textbf{\textsc{Generated Summary Bins}}} & \multirow{2}{*}{\textbf{\textsc{Sensitivity}}} \\
            & \textbf{\textsc{Length}} & \textbf{\textsc{1}} & \textbf{\textsc{2}} & \textbf{\textsc{3}} & \textbf{\textsc{4}} & \textbf{\textsc{5}} & \\
            \midrule
            \multirow{8}{*}{Llama3.1-8B-Instruct} & 1K & \textbf{0.76} & 0.75 & 0.69 & \underline{0.68} & \underline{0.68} & 4.0 \\
            & 2K & 0.73 & \textbf{0.76} & 0.73 & 0.70 & \underline{0.63} & 10.0 \\
            & 3K & 0.71 & 0.72 & 0.72 & \textbf{0.74} & \underline{0.62} & 10.25 \\
            & 4K & 0.71 & \textbf{0.75} & 0.70 & 0.70 & \underline{0.67} & 4.5 \\
            & 5K & 0.71 & 0.75 & \textbf{0.79} & \textbf{0.79} & \underline{0.70} & 6.0 \\
            & 6K & 0.69 & \textbf{0.76} & 0.73 & \textbf{0.76} & \underline{0.61} & 12.5 \\
\bottomrule
        \end{tabular}
    }
    \caption{Comparison of NLI model based faithfulness scores and Sensitivity for \textbf{long} summaries generated by different models across output bins on the Wikipedia dataset, with context lengths ranging from 1K to 8K. The highest faithfulness score in each bin is marked in \textbf{bold}, while the lowest is \underline{underlined}.}
\label{tab:bins_overall_models_long}
\end{table*}
\begin{table*}[h]
    \centering
    \resizebox{0.6\linewidth}{!}{
        \begin{tabular}{c|c|ccccc}
            \toprule
            \multirow{2}{*}{\textbf{\textsc{Models}}} & \textbf{\textsc{Context}} & \multicolumn{5}{c}{\textbf{\textsc{Generated Summary Bins}}} \\
            & \textbf{\textsc{Length}} & \textbf{\textsc{1}} & \textbf{\textsc{2}} & \textbf{\textsc{3}} & \textbf{\textsc{4}} & \textbf{\textsc{5}} \\
            \midrule
            \multirow{8}{*}{Llama3.1-8B-Instruct} & 1K & 0.24 & 0.29 & \underline{0.14} & 0.24 & \textbf{0.33} \\
            & 2K & 0.21 & 0.21 & \underline{0.16} & 0.24 & \textbf{0.37}  \\
            & 3K & 0.20 & 0.30 & 0.20 & \underline{0.10} & \textbf{0.57}  \\
            & 4K & \underline{0.06} & 0.17 & 0.23 & 0.17 & \textbf{0.26}  \\
            & 5K & 0.31 & \underline{0.20} & \underline{0.20} & \textbf{0.36} & 0.33 \\
            & 6K & 0.21 & 0.25 & 0.29 & \underline{0.14} & \textbf{0.50} \\
\bottomrule
        \end{tabular}
    }
    \caption{Comparison of FineSurE error rates for \textbf{standard} summaries generated by different models across output bins on the Wikipedia dataset, with context lengths ranging from 1K to 8K. The highest error rate in each bin is marked in \textbf{bold}, while the lowest is \underline{underlined}.}
\label{tab:bins_overall_models_short_finesure}
\end{table*}
\begin{table*}[h]
    \centering
    \resizebox{0.6\linewidth}{!}{
        \begin{tabular}{c|c|ccccc}
            \toprule
            \multirow{2}{*}{\textbf{\textsc{Models}}} & \textbf{\textsc{Context}} & \multicolumn{5}{c}{\textbf{\textsc{Generated Summary Bins}}} \\
            & \textbf{\textsc{Length}} & \textbf{\textsc{1}} & \textbf{\textsc{2}} & \textbf{\textsc{3}} & \textbf{\textsc{4}} & \textbf{\textsc{5}} \\
            \midrule
            \multirow{8}{*}{Llama3.1-8B-Instruct} & 1K & 0.53 & 0.27 & 0.60 & \underline{0.20} & \textbf{0.80} \\
            & 2K & 0.80 & \underline{0.53} & 0.67 & 0.80 & \textbf{1.47} \\
            & 3K & 1.75 & 1.12 & 1.12 & \underline{1.00} & \textbf{2.25} \\
            & 4K & 0.89 & 1.22 & 1.44 & \underline{0.44} & \textbf{2.44} \\
            & 5K & \underline{0.60} & 1.07 & 1.13 & 0.87 & \textbf{2.07} \\
            & 6K & 0.83 & \underline{0.67} & \underline{0.67} & \underline{0.67} & \textbf{1.50}  \\
\bottomrule
        \end{tabular}
    }
    \caption{Comparison of FineSurE error rates for \textbf{long} summaries generated by different models across output bins on the Wikipedia dataset, with context lengths ranging from 1K to 8K. The highest error rate in each bin is marked in \textbf{bold}, while the lowest is \underline{underlined}.}
\label{tab:bins_overall_models_long_finesure}
\end{table*}
\begin{table*}[h]
    \centering
    \resizebox{0.7\linewidth}{!}{
        \begin{tabular}{c|c|ccccc|c}
            \toprule
            \multirow{2}{*}{\textbf{\textsc{Models}}} & \textbf{\textsc{Context}} & \multicolumn{5}{c|}{\textbf{\textsc{Generated Summary Bins}}} & \multirow{2}{*}{\textbf{\textsc{Sensitivity}}} \\
            & \textbf{\textsc{Length}} & \textbf{\textsc{1}} & \textbf{\textsc{2}} & \textbf{\textsc{3}} & \textbf{\textsc{4}} & \textbf{\textsc{5}} & \\
            \midrule
            \multirow{8}{*}{Llama3.1-8B-Instruct} & 1K & \textbf{0.66} & 0.63 & 0.63 & 0.62 & \underline{0.57} & 6.5 \\
            & 2K & \textbf{0.70} & 0.62 & 0.64 & 0.60 & \underline{0.57} & 7.0 \\
            & 3K & \textbf{0.70} & 0.64 & 0.63 & 0.63 & \underline{0.61} & 4.0 \\
            & 4K & 0.60 & \textbf{0.66} & 0.63 & 0.61 & \underline{0.58} & 4.5 \\
            & 5K & \textbf{0.70} & 0.65 & 0.66 & 0.64 & \underline{0.61} & 5.25 \\
            & 6K & \textbf{0.70} & 0.67 & 0.69 & 0.66 & \underline{0.62} & 6.0 \\
\bottomrule
        \end{tabular}
    }
    \caption{Comparison of ROUGE-L based faithfulness scores and Sensitivity for \textbf{standard} summaries generated by different models across output bins on the Wikipedia dataset, with context lengths ranging from 1K to 8K. The highest faithfulness score in each bin is marked in \textbf{bold}, while the lowest is \underline{underlined}.}
\label{tab:bins_overall_models_short_rougel}
\end{table*}
\begin{table*}[h]
    \centering
    \resizebox{0.7\linewidth}{!}{
        \begin{tabular}{c|c|ccccc|c}
            \toprule
            \multirow{2}{*}{\textbf{\textsc{Models}}} & \textbf{\textsc{Context}} & \multicolumn{5}{c|}{\textbf{\textsc{Generated Summary Bins}}} & \multirow{2}{*}{\textbf{\textsc{Sensitivity}}} \\
            & \textbf{\textsc{Length}} & \textbf{\textsc{1}} & \textbf{\textsc{2}} & \textbf{\textsc{3}} & \textbf{\textsc{4}} & \textbf{\textsc{5}} & \\
            \midrule
            \multirow{8}{*}{Llama3.1-8B-Instruct} & 1K & \textbf{0.65} & 0.62 & 0.60 & 0.61 & \underline{0.58} & 4.0 \\
            & 2K & 0.65 & \textbf{0.66} & 0.64 & 0.62 & \underline{0.57} & 7.25 \\
            & 3K & 0.65 & \textbf{0.66} & 0.65 & 0.64 & \underline{0.56} & 9.0 \\
            & 4K & 0.62 & 0.61 & \underline{0.60} & \textbf{0.64} & \underline{0.60} & 1.75 \\
            & 5K & 0.65 & \textbf{0.66} & 0.65 & 0.63 & \underline{0.58} & 6.75 \\
            & 6K & 0.65 & \textbf{0.71} & \textbf{0.71} & 0.69 & \underline{0.61} & 8.0 \\
\bottomrule
        \end{tabular}
    }
    \caption{Comparison of ROUGE-L based faithfulness scores and Sensitivity for \textbf{long} summaries generated by different models across output bins on the Wikipedia dataset, with context lengths ranging from 1K to 8K. The highest faithfulness score in each bin is marked in \textbf{bold}, while the lowest is \underline{underlined}.}
\label{tab:bins_overall_models_long_rougel}
\end{table*}
\begin{table*}[h]
    \centering
    \small
    \resizebox{0.7\linewidth}{!}{
        \begin{tabular}{c|c|ccccc|c}
            \toprule
            \multirow{2}{*}{\textbf{\textsc{Datasets}}} & \textbf{\textsc{Context}} & \multicolumn{5}{c|}{\textbf{\textsc{Generated Summary Bins}}} & \multirow{2}{*}{\textbf{\textsc{Sensitivity}}} \\
            & \textbf{\textsc{Length}} & \textbf{\textsc{1}} & \textbf{\textsc{2}} & \textbf{\textsc{3}} & \textbf{\textsc{4}} & \textbf{\textsc{5}} & \\
            \midrule
            \multirow{8}{*}{Wikipedia}  & 1K & 0.81 & \underline{0.79} & \textbf{0.85} & 0.82 & \underline{0.79} & 2.75 \\
            & 2K & \textbf{0.84} & 0.79 & 0.81 & 0.81 & \underline{0.77} & 4.25 \\
            & 3K & \textbf{0.87} & 0.85 & 0.84 & 0.83 & \underline{0.79} & 5.75 \\
            & 4K & \textbf{0.91} & 0.90 & 0.88 & \underline{0.86} & 0.88 & 0.75 \\
            & 5K & 0.85 & 0.85 & \textbf{0.86} & \underline{0.83} & \underline{0.83} & 1.75 \\
            & 6K & \textbf{0.89} & 0.85 & 0.85 & 0.83 & \underline{0.81} & 4.5 \\
\bottomrule
        \end{tabular}
    }
    \caption{Comparison of MiniCheck-Flan-T5-Large based faithfulness scores and Sensitivity for \textbf{standard} summaries generated by the Llama3.1-8B-Instruct model across output bins in multiple domains, with context lengths ranging from 1K to 8K. The highest faithfulness score in each bin is marked in \textbf{bold}, while the lowest is \underline{underlined}.}
\label{tab:bins_overall_datasets_short_minicheck_t5}
\end{table*}
\begin{table*}[h]
    \centering
    \small
    \resizebox{0.7\linewidth}{!}{
        \begin{tabular}{c|c|ccccc|c}
            \toprule
            \multirow{2}{*}{\textbf{\textsc{Datasets}}} & \textbf{\textsc{Context}} & \multicolumn{5}{c|}{\textbf{\textsc{Generated Summary Bins}}} & \multirow{2}{*}{\textbf{\textsc{Sensitivity}}} \\
            & \textbf{\textsc{Length}} & \textbf{\textsc{1}} & \textbf{\textsc{2}} & \textbf{\textsc{3}} & \textbf{\textsc{4}} & \textbf{\textsc{5}} & \\
            \midrule
            \multirow{8}{*}{Wikipedia} & 1K & 0.78 & \textbf{0.82} & \underline{0.75} & \underline{0.75} & \underline{0.75} & 2.5 \\
            & 2K & \textbf{0.82} & 0.79 & 0.75 & 0.76 & \underline{0.73} & 5.0 \\
            & 3K & 0.77 & \textbf{0.81} & 0.80 & \textbf{0.81} & \underline{0.71} & 8.75 \\
            & 4K & 0.79 & \textbf{0.80} & \underline{0.73} & 0.77 & 0.74 & 3.25 \\
            & 5K & 0.80 & 0.83 & 0.83 & \textbf{0.84} & \underline{0.79} & 3.5 \\
            & 6K & 0.80 & 0.82 & 0.77 & \textbf{0.83} & \underline{0.72} & 8.5 \\
\bottomrule
        \end{tabular}
    }
    \caption{Comparison of MiniCheck-Flan-T5-Large based faithfulness scores and Sensitivity for \textbf{long} summaries generated by the Llama3.1-8B-Instruct model across output bins in multiple domains, with context lengths ranging from 1K to 8K. The highest faithfulness score in each bin is marked in \textbf{bold}, while the lowest is \underline{underlined}.}
\label{tab:bins_overall_datasets_long_minicheck_t5}
\end{table*}
\begin{table*}[h]
    \centering
    \small
    \resizebox{0.7\linewidth}{!}{
        \begin{tabular}{c|c|ccccc|c}
            \toprule
            \multirow{2}{*}{\textbf{\textsc{Datasets}}} & \textbf{\textsc{Context}} & \multicolumn{5}{c|}{\textbf{\textsc{Generated Summary Bins}}} & \multirow{2}{*}{\textbf{\textsc{Sensitivity}}} \\
            & \textbf{\textsc{Length}} & \textbf{\textsc{1}} & \textbf{\textsc{2}} & \textbf{\textsc{3}} & \textbf{\textsc{4}} & \textbf{\textsc{5}} & \\
            \midrule
            \multirow{8}{*}{Wikipedia} & 1K & 0.73 & 0.73 & \textbf{0.75} & 0.74 & \underline{0.71} & 3.0 \\
            & 2K & \textbf{0.77} & 0.76 & 0.75 & \underline{0.73} & \underline{0.73} & 2.25 \\
            & 3K & 0.79 & \textbf{0.80} & 0.76 & 0.76 & \underline{0.69} & 8.75 \\
            & 4K & 0.81 & \textbf{0.82} & 0.81 & 0.79 & \underline{0.76} & 4.75 \\
            & 5K & \textbf{0.80} & 0.79 & 0.79 & 0.76 & \underline{0.75} & 3.5 \\
            & 6K & \textbf{0.81} & 0.80 & 0.80 & 0.76 & \underline{0.71} & 8.25 \\
\bottomrule
        \end{tabular}
    }
    \caption{Comparison of NLI model based faithfulness scores and Sensitivity for \textbf{standard} summaries generated by the Llama3.1-8B-Instruct model across output bins in multiple domains, with context lengths ranging from 1K to 8K. The highest faithfulness score in each bin is marked in \textbf{bold}, while the lowest is \underline{underlined}.}
\label{tab:bins_overall_datasets_short}
\end{table*}
\begin{table*}[h]
    \centering
    \small
    \resizebox{0.7\linewidth}{!}{
        \begin{tabular}{c|c|ccccc|c}
            \toprule
            \multirow{2}{*}{\textbf{\textsc{Datasets}}} & \textbf{\textsc{Context}} & \multicolumn{5}{c|}{\textbf{\textsc{Generated Summary Bins}}} & \multirow{2}{*}{\textbf{\textsc{Sensitivity}}} \\
            & \textbf{\textsc{Length}} & \textbf{\textsc{1}} & \textbf{\textsc{2}} & \textbf{\textsc{3}} & \textbf{\textsc{4}} & \textbf{\textsc{5}} & \\
            \midrule
            \multirow{8}{*}{Wikipedia} & 1K & \textbf{0.76} & 0.75 & 0.69 & \underline{0.68} & \underline{0.68} & 4.0 \\
            & 2K & 0.73 & \textbf{0.76} & 0.73 & 0.70 & \underline{0.63} & 10.0 \\
            & 3K & 0.71 & 0.72 & 0.72 & \textbf{0.74} & \underline{0.62} & 10.25 \\
            & 4K & 0.71 & \textbf{0.75} & 0.70 & 0.70 & \underline{0.67} & 4.5 \\
            & 5K & 0.71 & 0.75 & \textbf{0.79} & \textbf{0.79} & \underline{0.70} & 6.0 \\
            & 6K & 0.69 & \textbf{0.76} & 0.73 & \textbf{0.76} & \underline{0.61} & 12.5 \\
\bottomrule
        \end{tabular}
    }
    \caption{Comparison of NLI model based faithfulness scores and Sensitivity for \textbf{long} summaries generated by the Llama3.1-8B-Instruct model across output bins in multiple domains, with context lengths ranging from 1K to 8K. The highest faithfulness score in each bin is marked in \textbf{bold}, while the lowest is \underline{underlined}.}
\label{tab:bins_overall_datasets_long}
\end{table*}

\end{document}